\documentclass[9pt,conference]{IEEEtran}
\IEEEoverridecommandlockouts

\usepackage{amsmath,amssymb,amsfonts}
\usepackage{algorithmic}
\usepackage{graphicx}
\usepackage{textcomp}

\usepackage{xcolor} 
\usepackage{verbatim}
\usepackage{xspace}
\usepackage{multirow}
\usepackage{array, makecell} %
\usepackage{subcaption}
\usepackage{booktabs}
\usepackage{nicefrac}
\usepackage{ulem} 
\usepackage{xcolor}
\usepackage{siunitx}
\definecolor{olivegreen}{rgb}{0, 0.6, 0}

\usepackage{pifont}
\newcommand{\cmark}{\color{olivegreen}\ding{51}}%
\newcommand{\xmark}{\color{red}\ding{55}}%

\def\BibTeX{{\rm B\kern-.05em{\sc i\kern-.025em b}\kern-.08em
    T\kern-.1667em\lower.7ex\hbox{E}\kern-.125emX}}
    
\usepackage[switch]{lineno}
    
\usepackage[backend=bibtex, sorting=none]{biblatex}
\AtBeginBibliography{\scriptsize}
\bibliography{refs}
\begin{document}

\newcommand{\expnumber}[2]{{#1}\mathrm{e}{#2}}

\newcommand{\JL}[1]{{\color{magenta}[\textbf{\sc JLee}: \textit{#1}]}}

\newcommand{\JLr}[1]{{\color{magenta}[\textbf{\sc JLee}: \textit{#1}]}}

\newcommand{\JS}[1]{{\color{blue}[\textbf{\sc from JS}: \textit{#1}]}}

\newcommand{\HJ}[1]{{\color{green}[\textbf{\sc from HYoon}: \textit{#1}]}}
\newcommand{\rev}[1]{{\color{olivegreen}[{#1}]}}
\renewcommand{\rev}[1]{#1}

\newcommand{\deltgt}[1]{\sout{\color{magenta}[{#1}]}}

\newcommand{\nasname}{DANCE\xspace}

\title{DANCE: Differentiable Accelerator/Network Co-Exploration}

\author{%
\IEEEauthorblockN{
Kanghyun Choi\textsuperscript{1}\textsuperscript{$\dagger$}, 
Deokki Hong\textsuperscript{2}\textsuperscript{$\dagger$}, 
Hojae Yoon\textsuperscript{1}\textsuperscript{$\dagger$}, 
Joonsang Yu\textsuperscript{3}, 
Youngsok Kim\textsuperscript{1}\textsuperscript{2}, 
and Jinho Lee\textsuperscript{1}\textsuperscript{2}\textsuperscript{*}
}%
\IEEEauthorblockA{\textsuperscript{1}Department of Computer Science, Yonsei University}%
\IEEEauthorblockA{\textsuperscript{2}Department of Artificial Intelligence, Yonsei University}%
\IEEEauthorblockA{\textsuperscript{3}Department of Electrical and Computer Engineering, Seoul National University}%
\textsuperscript{1,}\textsuperscript{2}\{kanghyun.choi, dk.hong, johnyoon, youngsok, leejinho\}@yonsei.ac.kr \ \ 
\textsuperscript{3}shorm21@dal.snu.ac.kr%
\thanks{Author preprint. Accepted to DAC'21}\\
\thanks{\textsuperscript{$\dagger$}Equal contribution.}
\thanks{\textsuperscript{*}Correspondence.}
}

\maketitle

\begin{abstract}
This work presents \nasname, a differentiable approach towards the co-exploration of hardware accelerator and network architecture design. At the heart of \nasname is a differentiable evaluator network. By modeling the hardware evaluation software with a neural network, the relation between the accelerator design and the hardware metrics becomes differentiable, allowing the search to be performed with backpropagation. Compared to the naive existing approaches, our method performs co-exploration in a significantly shorter time, while achieving superior accuracy and hardware cost metrics.

\end{abstract}



\normalem 

\section{Introduction}

After decades of efforts from researchers, DNNs now exhibit near- or over-human performance in various applications domains such as image classification and playing board games~\cite{alphagozero}. 
However, the success comes at the cost of exploding compute intensity, 
which leads to long training GPU hours and large hardware costs.

\emph{Neural Architecture Search} (NAS) is an approach that tries to solve such problems. 
It has started with the aim of reducing the human design effort and achieving state-of-the-art accuracy~\cite{darts, mnasnet}, but now starting to consider hardware-related costs such as latency~\cite{proxylessnas}.

Another popular method to tackle the problem is through specialized hardwares (Often called \emph{`accelerator's}. In this paper, we use words `hardware' and `accelerator' interchangeably). 
By utilizing an accelerator specialized for executing DNNs, they achieve superior latency and/or cost~\cite{diannao, eyeriss, tpu}. 
For example, Google TPU~\cite{tpu} has been deployed to accelerate the processing of AlphaGo~\cite{alphagozero}, datacenters and cloud services. 
Designing dedicated accelerators opens up another large design space for optimizing not only latency, but also other hardware cost metrics such as energy consumption and area. 

However, network architectures and accelerators are not independent, and blindly optimizing one side could often hurt another.
For example, the commonly used separable convolution~\cite{mobilenet} usually achieves superior latency due to its low computational requirements.
However, some type of accelerators such as Google's TPU are designed to take advantage of large number of output channels for parallelism.
Because of this, a separable convolution executed on TPU suffers from long latency compared to normal convolution operations despite the less number of computations~\cite{edgetpunas}. 
Similarly, optimizing only the accelerator without considering the network would often yield suboptimal solutions.

In such regard, co-exploration of both the hardware accelerator and network architectures~\cite{cosearch, cosearch_fpga, cosearch_fpga2, cosearch_multi, cosearch_bestofboth} is critical in achieving both the desired application performance (i.e., accuracy) and a reasonable cost (latency, area, and energy consumption). 
Existing co-exploration techniques typically achieve the goal using Reinforcement Learning (\emph{RL}) techniques. 
They first generate a network/accelerator pair, which is evaluated by training the network for accuracy and measuring the hardware cost metrics. 
After evaluation, a reward function is calculated and a new design pair is generated based on the reward.
The obvious problem with this procedure is that it requires a huge search time.
As in RL-based NAS techniques, the generated network needs to be fully trained in order to evaluate the accuracy. 
Also, the accelerator evaluation often takes non-negligible time and resources to complete.
Hence, the search demands excessive amount of time and is still difficult to achieve high-quality solutions~\cite{cosearch, cosearch_bestofboth}. 

\begin{figure*}[t]
    \centering
    \subcaptionbox{Seven dimensions of conv layers.\label{fig:cnn}}{
        \centering
        \includegraphics[width=0.23\textwidth]{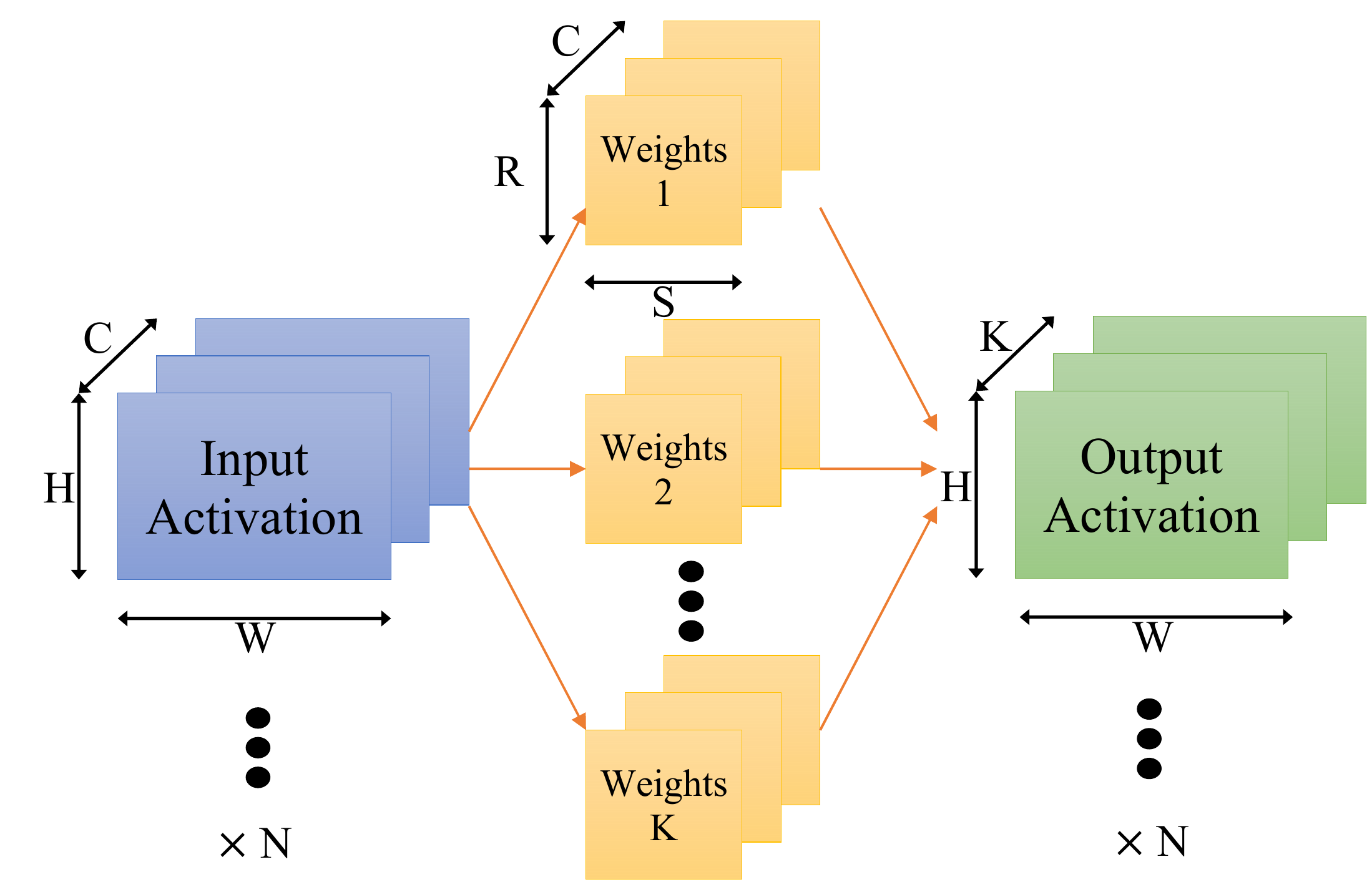} 
    }
    \subcaptionbox{CNN execution.\label{fig:cnncode}}{
        \centering
        \includegraphics[width=0.24\textwidth]{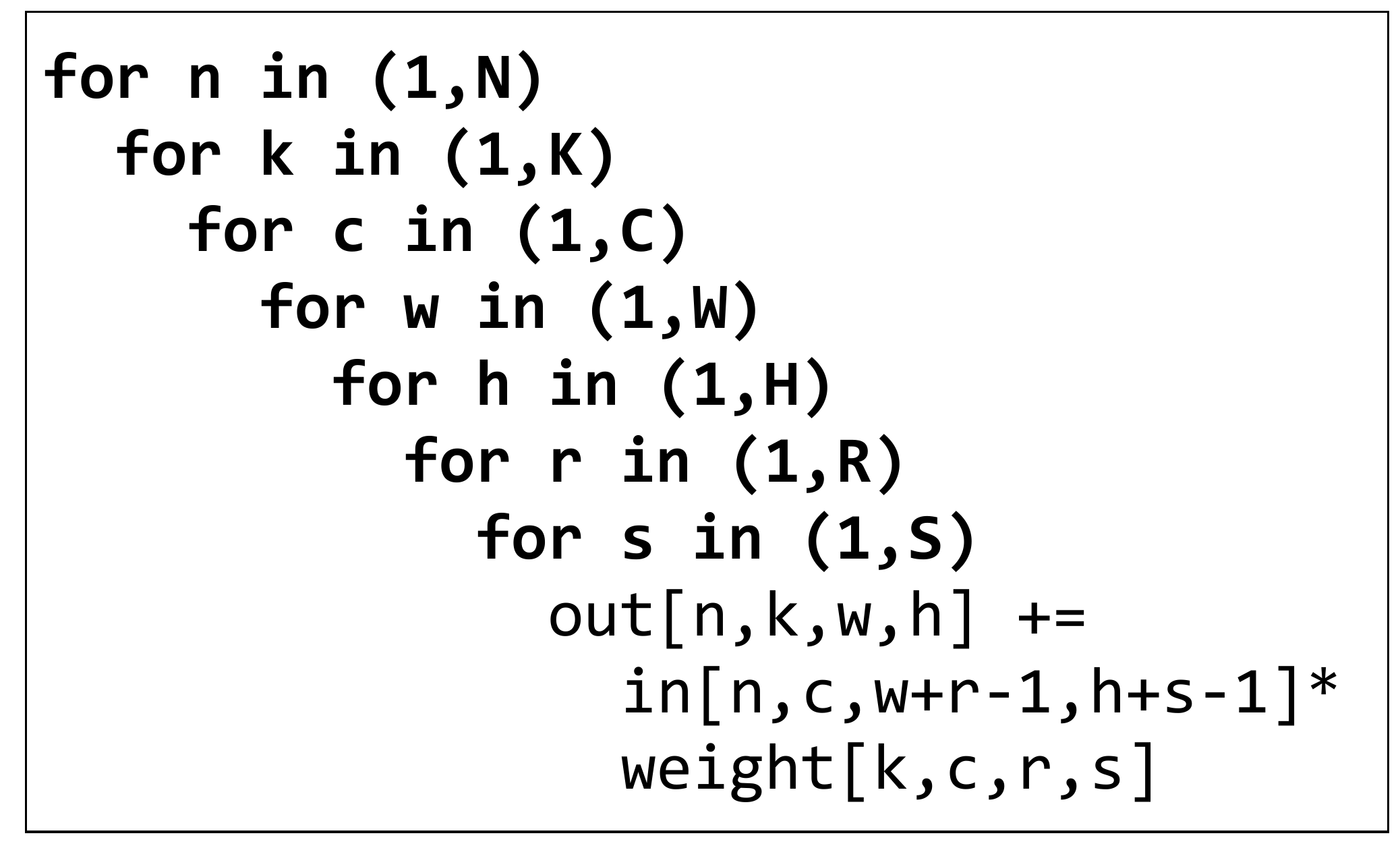} 
    }    
    \subcaptionbox{An example DNN accelerator.\label{fig:eyeriss}}{
        \centering
        \includegraphics[width=0.3659\textwidth]{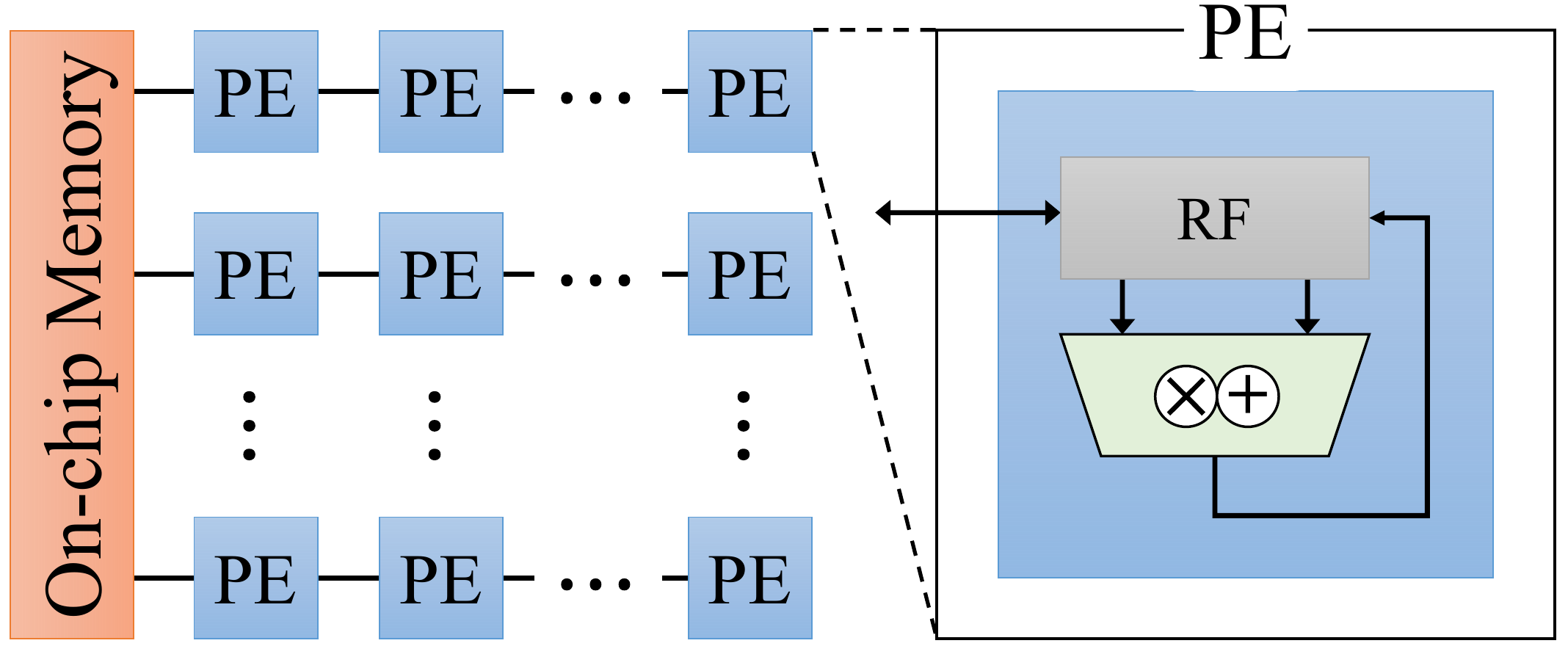} 
    }
    \caption{CNN execution dimensions and an example DNN accelerator. (a) and (b) shows the  seven dimensions in a convolutional layer and its execution, and (c) shows an example DNN accelerator with multiple PEs (Processing Elements).}\vspace{-3mm}
    \label{fig:hardware}
\end{figure*}

This work presents \emph{\nasname} (Differentiable Accelerator/Network Co-Exploration), 
which hugely reduces the search cost by adopting the idea of differentiable neural architecture search~\cite{darts} into the co-exploration problem. 
Differentiable NAS reduces the search cost of the NAS problem by modeling the network architecture space as a continuous hyperspace.
However, the search cannot be directly applied to accelerator/network co-exploration, as the relation between the accelerator design and the hardware metrics is discrete and complicated~\cite{timeloop, maestro}.
\rev{A previous work~\cite{edd} tries to formulate a differentiable co-exploration in a limited way, but models the execution time only with flops per resource, hindering the true relation between network architectures and accelerator designs.} 

At the heart of \nasname is the modeling of the accelerator evaluation software (which is non-differentiable) using a neural network that
can be used as a differentiable loss function. 
Using the evaluator network, we propose a method for exploring neural architectures and accelerator designs.

\nasname can be applied to any differentiable NAS framework, using any evaluation software such as simulators~\cite{scalesim} or schedulers~\cite{timeloop, maestro}.
We demonstrate that \nasname can efficiently explore the search space within a short amount of time, and discover good design points among accuracy, latency, energy efficiency and area. Our contributions can be summarized as the following:
\begin{itemize}
    \item We propose a differentiable co-exploration method for accelerators and network architectures to achieve fast search time and application performance at the same time. 
    \item We provide a cost estimation network architecture for modeling the accelerator evaluation which is differentiable and fast, with high accuracies. \rev{To the best of our knowledge, this is the first work that models the relation between network architectures and hardware accelerator designs in a differentiable way}.
    \item We provide a hardware generation network architecture that searches for the optimal accelerator design given a network design under search.
\end{itemize}


\section{Related Work}
\subsection{Neural Architecture Search}
In reaction to cope with the increasing network size and the corresponding manual design effort, neural architecture search automates the design of DNN architectures. 
Early works usually adopted RL or EA to generate the network~\cite{mnasnet, amoebanet}.

However, the search cost is huge for those algorithms, with up to thousands of GPU-days due to the full training required for every candidate.
Differentiable neural architecture search~\cite{darts} is a way to mitigate such cost, which builds a supergraph and finds a path within it. 
It was able to find state-of-the-art performing networks within a few orders of magnitude shorter time.

Some recent works started employing hardware penalties on the differentiable NAS. 
ProxylessNAS~\cite{proxylessnas} and FBNet~\cite{fbnet} add latency constraints to NAS with lookup tables.
LA-DARTs~\cite{la-darts} and AOWS~\cite{aows} adopt ML-based modeling for network latency. 
Finally, OFA~\cite{ofa} is a work worth mentioning, that searches a supernetwork that can be later cropped to adapt towards various cost requirements.

\subsection{Hardware DNN Accelerators}
\label{sec:hardware}
Hardware accelerators for DNNs are often focused on parallely executing multiple Multiply-Accumulate (MAC) operations which are the most common operation in modern CNNs. 
%
\figurename~\ref{fig:eyeriss} shows an example Eyeriss~\cite{eyeriss}-like DNN accelerator that comprises on-chip memory, many PEs (processing elements), and the interconnects between them.
Even with a backbone accelerator design, many properties still remain to be designed, such as the number of PEs, dataflow, and register file size, etc.

Usually, a DNN layer includes multiple dimensions of computations. 
For example, a convolution layer contains seven layers of computations as shown in \figurename~\ref{fig:cnn}: three for input activations ($H$, $W$, $C$), three for weights ($R$, $S$, $K$), and one for batches ($N$). 
Thus, it is formulated as a seven-level nested loop (\figurename~\ref{fig:cnncode}). 
Mapping and ordering of those loops on accelerators
is often called \emph{dataflow}, and many accelerators \cite{eyeriss, shidiannao} provide different dataflows that focus on keeping some of the data on local memory as long as possible.

Analyzing how each choice in the accelerator design affects the DNN latency is often performed by simulators~\cite{scalesim} or analytical evaluation tools~\cite{timeloop, maestro}. 
In this work, we utilize Timeloop~\cite{timeloop} combined with Accelergy~\cite{accelergy}, a state-of-the-art accelerator evaluation toolchain to train the evaluation network within the \nasname framework.

\begin{figure}[t]
\centering

    \includegraphics[width=\columnwidth]{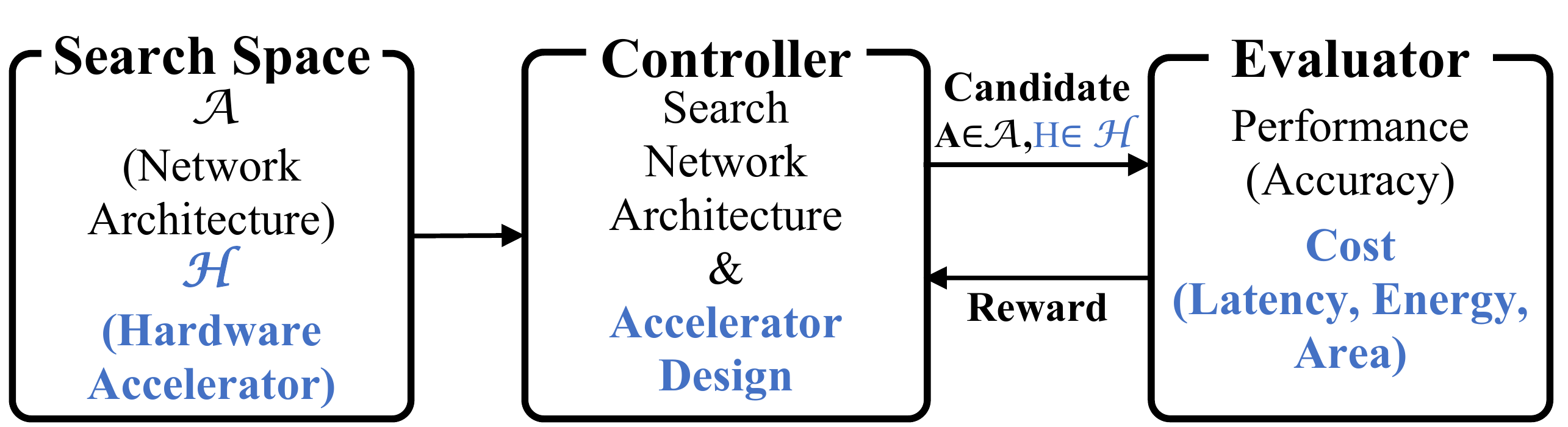}
    \caption{RL-based co-exploration\label{fig:nas-RL}. The black letters represent the parts present for RL-based NAS algorithms, and the blue letters represent the parts added for co-exploration.}\vspace{-3mm}
\label{fig:rl}
\end{figure}

\subsection{Accelerator/Network Co-exploration}
There are several existing works on co-exploring the network architecture and accelerator design. 
Most of the prior work on co-exploration utilize RL as their controller due to its relatively simpler way of formulating the problem. 
\cite{cosearch,cosearch_bestofboth, cosearch_multi, cosearch_fpga2} propose such RL-based algorithms for co-exploring the accelerators and network architectures under various environments. 
However, they all inherit the same search-cost problem from the RL-based NAS algorithms. 

In contrast, this paper proposes adopting the idea of differentiable NAS on the co-exploration problem to significantly reduce the search cost, while still producing state-of-the-art accuracy network and accelerator design.
A previous work named EDD~\cite{edd} provides a differentiable method for the co-exploration problem. However, the work has a few significant limitations. 
Most importantly, EDD models the latency as the total flops of the network divided by the amount of computation resource. 
As a result, the true relation between the network architecture and the accelerator design is not considered in the co-search. 
This theoretically does not allow searching for a few critical features such as dataflows or register file sizes.
Also, the main focus of EDD is to use varying quantization for each layer. 
Thus, the work assumes that there are dedicated hardware for each layer (with possible sharing), and is far from general accelerators~\cite{tpu, eyeriss, shidiannao}.
We provide a discussion with a quantitative comparison in Section~\ref{sec:eval}.
A non peer-reviewed concurrent work~\cite{dna}
creates a table of candidate accelerator designs, and selects them using Gumbel softmax. 
While it solves a similar problem with ours, it has a problem of having to evaluate all the hardware candidate designs against all the layers of the network beforehand and is not scalable.

\section{Differentiable Accelerator/Network Co-Exploration}
\subsection{Existing RL-Based Co-explorations}
\figurename~\ref{fig:nas-RL} shows how RL-based co-explorations are performed.
The black letters represent the components for ordinary NAS algorithms, and the blue letters represent components added for the co-exploration.
First, the search space on the network architecture and hardware accelerator is provided to the controller.
Then the controller generates candidate designs for the provided space (network and accelerator).
The generated candidates are sent to the evaluator, which performs training on the network to get its accuracy, and analyzes the cost metrics for the given hardware running the network.
While it well-serves the purpose of co-exploration, it shares the same problem of RL-based NAS algorithms: the training cost. 
A costly training is required for each candidate being generated. 
Moreover, searching for the optimal hardware design also takes considerable time which is also performed for each candidate. 
As a result, the search suffers from exploding GPU-hours.

\begin{figure}
    \includegraphics[width=\columnwidth]{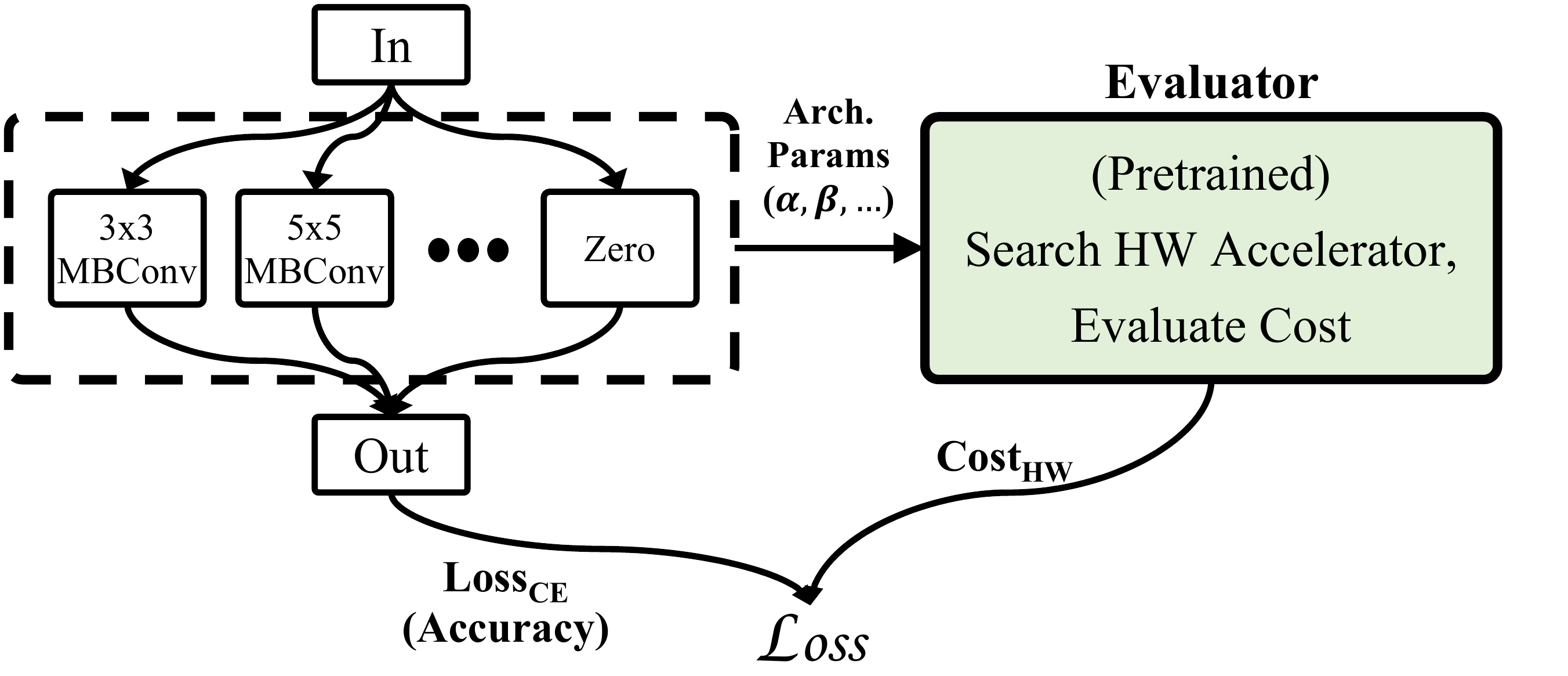}
          \caption{\nasname with differentiable co-exploration.}
          \label{fig:nas-diff}\vspace{-3mm}
\end{figure}

\subsection{Proposed Differentiable Co-exloration}
\figurename~\ref{fig:nas-diff} illustrates the proposed differentiable co-exploration method, named \nasname.
The left part is the network search module similar to other differentiable NAS algorithms, where a path within the super-network is found using backpropagation to become the final searched network.
In fact, any differentiable NAS algorithm can be used, and is orthogonal to the proposed approach.
On the right-hand side of \figurename~\ref{fig:nas-diff} is the differentiable evaluator, which takes the architecture parameters from the search module, searches for the optimal hardware accelerator design, and evaluates its cost metrics.
It is a pre-trained neural network, which is frozen during search and used only to connect the architecture to the hardware cost metrics.
The loss function considers both the accuracy and the cost metrics as below:
\vspace{-3mm}
\begin{equation}
    \mathcal{L}oss = \mathcal{L}oss_{CE} + \lambda_1||w|| + \lambda_2Cost_{HW}
    \label{eq:overall}
\vspace{-0.2mm}
\end{equation}
Where  $\lambda_1$ and $\lambda_2$ are hyperparameters that control the trade-off between the terms.
$\mathcal{L}oss_{CE}$ is the cross-entropy loss, and $||w||$ is the weight decay term following \cite{proxylessnas}.
Finally, $Cost_{HW}$ is the cost function for hardware accelerators, calculated from the output values of the evaluator network. 
For example, it can be a linear combination of latency, area, and the energy consumption, or EDAP (Energy-delay-area product).
We discuss various possible hardware cost functions in Section~\ref{sec:cost}.

\subsection{Evaluator Network}
\label{sec:cascaded}
The original (non-differentiable) cost evaluation softwares are comprised of a hardware generation tool and a cost estimation tool.
As briefly explained in Section~\ref{sec:hardware}, the hardware generation tool takes the network architecture as the input, and proposes a hardware accelerator design. 
In this paper, we choose dataflow, number of PEs along the X and Y dimension, and size of the register file as the search space of the hardware accelerator design (Refer to Section~\ref{sec:space} for more details).
Then, the cost estimation tool takes the hardware accelerator and network architecture to output the cost metrics. 
In general, the hardware generation tool is composed as an outer loop enclosing the cost estimation tool. 
By using exact algorithms such as exhaustive search or branch-and-bound algorithms, it outputs the optimal solution for the given network architecture, within the hardware search space $\mathcal{H}$.
In this work, we use Timeloop~\cite{timeloop} for latency and Accelergy~\cite{accelergy} for energy/area, which is regarded as a state-of-the-art cost estimation toolchain.
On top of those, we design our own hardware generation tool.
We generate random networks within the network architecture space $\mathcal{A}$ as inputs, and the output of the toolchain will become ground-truth for training the components for evaluator network.

\begin{figure}
\centering
    \includegraphics[width=0.9\columnwidth]{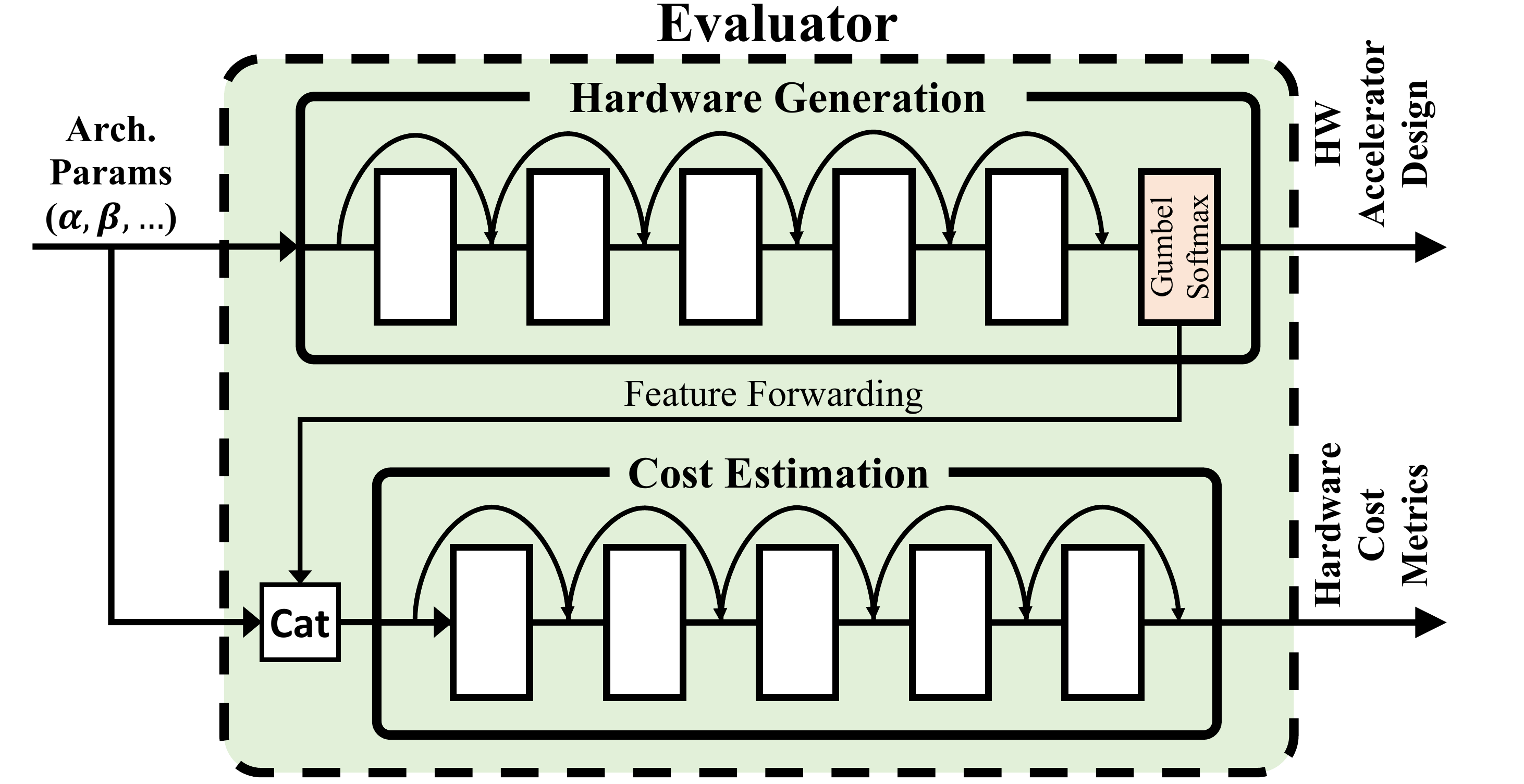}
          \caption{Evaluator network architecture. It is composed of a hardware generation network and a cost estimation network. \vspace{-1mm}
          }
          \label{fig:evaluator}
\end{figure}

Our design of the evaluator network is composed of two modules: Hardware generation network and cost estimation network. 
\figurename~\ref{fig:evaluator} shows the evaluator network architecture.
We model the hardware generation with a five-layer perceptron, which uses ReLUs as activation functions. 
To increase the accuracy of the cost estimation network and establish the gradient path towards the network under search, we adopt residual connections between the layers~\cite{resnet}.

For the cost estimation, we model the network as a five-layer regression with residual connections. 
It has ReLU as activation functions, and applies batch normalization every layer. 
It outputs the three cost metrics of our interest (latency, area, and energy consumption), based on the ground truth generated from the evaluation software (Timeloop + Accelergy) described above.
We use MSRE (Mean Squared Relative Error) loss for training each evaluator network, which can be represented as the following:
\vspace{-1mm}
\begin{equation}
    \mathcal{L}oss_{MSRE} = \textstyle\sum_i{(1 - \nicefrac{\hat{y_i}}{y_i})^2}
\vspace{-1mm}
\end{equation}
where $y_i$ is the hardware cost function ($Cost_{HW}$) for each metric generated from result of Timeloop+Accelergy and $\hat{y_i}$ is the same cost function calculated using the network output. 
While we could use typical MSE loss, it has a problem of falsely heavy-weighing the high value metrics.
For example, the latency values outputted within our search space ranges from 8ns to over 100ns per each layer.
If we use MSE loss, we regard the 10ns error out of 8ns latency and 10ns error out of 100ns latency as the same, giving unfair favor towards modeling the long-latency situations more correct. 
Since our objective is to find accelerators with low latency, MSRE loss is more desired. 

In the evaluator architecture, the cost estimation network outputting the HW cost metrics means that it has to internally model two functions: Find the optimal hardware, and estimate the metrics. 
While a standalone network shows fairly high accuracy (Section~\ref{sec:eval}), we can further improve the latency by adding a feature forwarding path from the output of the hardware generation network. 
We concatenate the result of the hardware generation network to the network architecture as an input to the cost estimation network.
We use Gumbel softmax~\cite{gumbel} as the last layer of the hardware generation network, so that the output value from the hardware generation becomes as close as possible to the input of the cost estimation network.

\subsection{Hyperparameter Warm-up}
\label{sec:warmup}
Compared to optimizing for the application's classification accuracy, optimizing for the cost metrics is relatively easier for gradient descent.
For instance, selecting most layers to be zero quickly optimizes all of the latency, area, and energy consumption. 
Once the network architecture falls into such a solution, it is difficult to find heavier architectures even if those are needed for optimizing the best accuracy.
To mitigate such effect, we use hyperparameter warm-up scheduling. 
We use small $\lambda_2$ from Eq. \ref{eq:overall} for the first few epochs, and increase the $\lambda_2$ to the desired value later, after the network architecture reaches a certain stage for high accuracy.

\subsection{Hardware Cost Functions}
\label{sec:cost}
By default, we use a linear combination of the three hardware cost metrics as the cost function $Cost_{HW}$ of Eq. \ref{eq:overall} as below:
\vspace{-1mm}
\begin{equation}
\small
    Cost_{HW\_linear} = \lambda_EEnergy + \lambda_LLatency + \lambda_AArea
\vspace{-1mm}
\end{equation}
By controlling $ \lambda_E$, $\lambda_L$ and $\lambda_A$, we can set a constraint on how we weigh the balance between each cost metric. 
\rev{We use \SI{}{\milli\joule}, \SI{}{\milli\second} and \SI{}{\micro\meter\squared} units for each cost to match the scale of these hyperparameters.}
In addition, we use the product of all metrics, 
\vspace{-1mm}
\begin{equation}
\small
    Cost_{HW\_EDAP} = Energy\cdot Latency \cdot Area
\vspace{-1mm}
\end{equation}
where EDAP is a common metric used to evaluate hardwares (i.e. \emph{energy-delay-area product}.
It presents the benefit of having no extra hyperparameter and is unitless. 
Refer to Section~\ref{sec:eval} for how setting those cost functions affect the searched solutions.


\section{Experimental Results}
\label{sec:eval}
We have conducted several experiments on \nasname using CIFAR-10 and ImageNet (ILSVRC2012) dataset. 
In this section, we describe the results of the evaluator network, and then the co-exploration solution quality in comparison with a few prior works. 
All algorithms are implemented on PyTorch 
and run on four RTX2080Ti GPUs.

\begin{table}[]
    \centering
    \scriptsize
    
    \caption{Performance of the Evaluator Network}
    \label{tbl:evaluator}
    
    \begin{tabular}{ccccc}
    \toprule
    Network & \multicolumn{4}{c}{Accuracy}\\
    \midrule
    \multirow{2}{*}{Hardware Generation} & $PE_X$ & $PE_Y$ & RF\_Size & Dataflow \\
     & 98.9\% & 98.3\%  & 98.3\% & 98.8\% \\
    \midrule
    
    \multirow{2}{*}{\makecell{Cost Estimation \\ (w/o feature forwarding)}} &  Latency &  Energy & Area \\
     & 93.7\% & 96.3\% & 92.8\% \\
     \midrule
     \multirow{2}{*}{\makecell{Cost Estimation \\ (w/ feature forwarding)}}   & Latency &  Energy & Area \\

     & 99.6\%  & 99.7\% & 99.9\% \\
    \midrule

    \multirow{2}{*}{Overall Evaluator} & Latency &  Energy & Area \\
     & 98.3\% & 98.3\% & 99.2\% \\
    \bottomrule
    \end{tabular}
    \vspace{-3mm}
\end{table}

\begin{table*}[]
    \centering
    \scriptsize
    
    \caption{Performance of DANCE on CIFAR-10}
    \label{tbl:cifar}
    
    \def\arraystretch{0.9}%
    \begin{tabular}{cp{6mm}p{6mm}p{6mm}clccccc}
    \toprule
    \multirow{2}{*}{$HW\_{Cost}$} & \multicolumn{3}{c}{Cost Hyperparam.} &\multirow{2}{*}{$\lambda_2$} & \multirow{2}{*}{~~~~~~~~~~Method} & Acc. & Latency & Energy & EDAP & Dataflow \\
    &$\lambda_L$&$\lambda_E$&$\lambda_A$&&&  (\%) & (\SI{}{\milli\second}) & (\SI{}{\milli\joule}) & \tiny(\SI{}{\joule} $\cdot$\SI{}{\second} $\cdot$\SI{}{\meter\squared} $\cdot 10^{-18}$ )\\ 
      \midrule
      \multirow{6}{*}{$Cost_{HW\_EDAP}$} &  \multirow{6}{*}{N/A} &  \multirow{6}{*}{N/A} &  \multirow{6}{*}{N/A}  &     
           N/A            & Baseline (No penalty) + HW$^\dagger$    &   95.08 & 11.0 & 22.1 & 440.4 &   OS \\  
      &&&& $\expnumber{3.33}{-9}$ & Baseline (Flops penalty) + HW-A &   94.97 &  9.3 & 18.2 & 421.1 & OS  \\ 
      &&&& $\expnumber{1.03}{-8}$ & Baseline (Flops penalty) + HW-B &   93.10 &  5.7 & 22.4 & 321.2 & WS  \\ 
      &&&& $\expnumber{1.39}{-9}$ & EDD (Original)                  &   83.59 &	 0.8 &	1.0 &   2.1 & RS  \\ 
      &&&& $\expnumber{3.33}{-7}$ & EDD + Proposed Loss func.-A           &   94.48 &	 7.4 & 14.5 & 218.7 & OS  \\ 
      &&&& $\expnumber{2.78}{-7}$ & EDD + Proposed Loss func.-B           &   93.90 &  6.9 &  5.3 &  74.3 & RS  \\ 
     \midrule
      \multirow{2}{*}{$Cost_{HW\_EDAP}$} &  \multirow{2}{*}{N/A} &  \multirow{2}{*}{N/A} &  \multirow{2}{*}{N/A} & 
          $\expnumber{2.00}{-4}$  & \nasname-A                    & 94.50   &  5.6 &  5.6 & 70.0  & RS \\  
      &&&&$\expnumber{6.67}{-4}$  & \nasname-B                    & 93.25   &  3.2 &  5.3 & 30.9  & OS \\ 
  
      \midrule
      
      \multirow{2}{*}{\makecell {$Cost_{HW\_linear}$ \\ (Latency-Oriented)} } & \multirow{2}{*}{3.3} &\multirow{2}{*}{0.8} &\multirow{2}{*}{1.0}& 
          $\expnumber{1.43}{-3}$  & \nasname-A                    & 94.65   & 5.9  &  7.5 & 119.1 & RS\\ 
      &&&&$\expnumber{6.67}{-3}$  & \nasname-B                    & 93.23   & 1.8  &  7.9 &  36.9 & WS \\ 
      
    \midrule
                  
      \multirow{2}{*}{\makecell {$Cost_{HW\_linear}$ \\ (Energy-Oriented)} }  &  \multirow{2}{*}{0.2} &\multirow{2}{*}{2.8} &\multirow{2}{*}{1.0} & 
          $\expnumber{1.67}{-3}$  & \nasname-A                    & 94.98   & 7.5  &  7.8 & 132.5 & RS  \\ 
      &&&&$\expnumber{6.67}{-3}$  & \nasname-B                    & 94.00   & 4.6  &  3.7 &  31.1 & RS\\ 
       
    \midrule
    
      \multirow{2}{*}{\makecell {$Cost_{HW\_linear}$ \\ (Balanced)}}& \multirow{2}{*}{0.6} &\multirow{2}{*}{0.5} &\multirow{2}{*}{1.0} & 
          $\expnumber{5.00}{-3}$ & \nasname-A                     & 94.71   & 5.5  &  6.9 & 103.2 & RS    \\   
      & && &  
      $\expnumber{8.33}{-3}$ & \nasname-B & 93.73 & 4.2 & 4.4 & 41.5 & RS \\ 

              \bottomrule
        \multicolumn{11}{r}{\tiny$^\dagger$HW: Hardware accelerator, designed using the hardware generation tool after the NAS.} \\
    \end{tabular}
    \vspace{-5mm}
\end{table*}

\subsection{Search Space}
\label{sec:space}
For $\mathcal{H}$, the hardware accelerator search space, we use a state-of-the-art accelerator Eyeriss~\cite{eyeriss} as the backbone. 
We choose number of PEs, RF size, and Dataflow as design parameters.
For the two-dimensional PE array, we separately assign a variable per dimension: $PE_X$ and $PE_Y$, where each value can range from 8 to 24.
In our settings, larger $PE_X$ favors the layers with more channels, and larger $PE_Y$ favors larger feature maps for parallelism.
The RF size per PE could take values between 4 and 64.
For Dataflow, we choose three dataflows from existing hardware accelerators: WS (Weight Stationary~\cite{tpu}), OS (Output Stationary~\cite{shidiannao}), and RS (Row Stationary~\cite{eyeriss}).
We assumed a HBM memory with about 128GB/s for the off-chip memory.
Within the evaluator network, each variable is formulated as one-hot vectors to simplify the cascaded connection between the hardware generation and the cost estimation networks.

For the network architecture search space $\mathcal{A}$, we have adopted ProxylessNAS~\cite{proxylessnas} as the backbone network architecture. 
The network has 13 layers, where the number of channels increase every 3 layers. 
Each of the 9 layers placed in the middle has 7 candidate operations in addition to a skip connection: {MBConv3X3\_expand3, MBConv3X3\_expand6, MBConv5X5\_expand3, MBConv5X5\_expand6, MBConv7X7\_expand3, MBConv7X7\_expand6, and Zero}. 
When Zero is chosen, only the skip connection remains and the layer effectively disappears from the network.
The architecture parameters are trained using the binarized method as in~\cite{proxylessnas}.
Note that \nasname is orthogonal to the network architecture search space and algorithm, and we choose \cite{proxylessnas} as it is a popular algorithm for NAS.

\subsection{Evaluator Network Results}

\subsubsection{Cost Estimation Network}
\tablename~\ref{tbl:evaluator} shows the experimental results for the components of the evaluator network. We separately train the cost estimation and the hardware generation network on the ground truth values and combine them. 

Each layer of the cost estimation network has width of 256, 
and the network is trained using Adam optimizer with learning rate of 0.0001 for 200 epoches.
We have used batch size of 256.
We have trained the cost estimation network on 1.8 million cases generated with Timeloop+Accelergy from the search space, and validated on 0.45 million cases.
All three cost metrics show over 99\% accuracy, showing that it is precise enough.
 Also, it is observed that feature forwarding improves the accuracy by 4.3\%p on average.

\subsubsection{Hardware Generation Network}
For the hardware generation network, 
the layer widths are set to be 128.
We use normal CE loss as the loss function, 
which we denote as $\mathcal{L}oss_{CE\_HW}$.
The network has been trained using SGD with batchsize 128 for 200 epoches, where the learning rate starts with 0.001 and decreases by $0.1\times$ every 50 epoches. 
We have generated 50K network cases from the search space explained in the previous section, and used 10K cases for validation.
On all of the hardware accelerator design parameters, the accuracy was nearly 99\%, also showing that it is accurate enough.
It is worth noting that the hardware generation network is not only accurate and differentiable, but it is also much faster than the original generation toolchain. 
With the same functionality, the inference time for the hardware generation network takes about 0.5ms with a single GPU, while the generation tool takes about 112s using 48 threads from 24 cores of two Intel Xeon Silver-4214 CPUs.

\subsubsection{End-to-end Evaluator Network Results}
Finally, we test the whole evaluator network as a combination of hardware generation and cost estimation network. 
Even though the intermediate values are not one-hot vectors, Gumbel softmax approximates them well, and still maintains around 99\% accuracy for the cost metrics.

\begin{table}[b]
\vspace{-1mm}
    \centering
    
\scriptsize
    \caption{Performance of DANCE on ImageNet}
    \label{tbl:imgnet}
    
    \begin{tabular}{lcccc}
    \toprule
     Method  & Acc. & Latency & Energy & EDAP \\
    \midrule
        Baseline (No penalty) + HW     & 71.12\% & 23.3ms &  71.6mJ & 3014.0 \\
        Baseline (Flops Penalty) + HW  & 70.56\% & 13.4ms &  70.9mJ & 2709.0 \\
        EDD + Proposed Loss func.      & 70.34\% & 28.1ms &  94.8mJ & 5642.5 \\
        \nasname ($Cost_{HW\_EDAP}$)   & 69.82\% &  7.5ms &  42.7mJ &  912.4 \\
        \nasname (Energy-Oriented)     & 69.55\% &  9.2ms &  49.5mJ & 1413.5 \\
        \nasname (Latency-Oriented)    & 70.41\% &  8.3ms &  48.4mJ & 1154.3\\
        \nasname (Balanced)            & 70.15\% &  7.7ms &  45.7mJ & 1001.8 \\
\bottomrule
    \end{tabular}
\end{table}

\subsection{Co-exploration Results}
\label{sec:co-ex}

\subsubsection{Experimental Results on CIFAR-10}
\tablename~\ref{tbl:cifar} shows the performance of \nasname on CIFAR-10 dataset.
For the first baseline, we have performed search using ProxylessNAS~\cite{proxylessnas} (w/ or w/o Flops penalty term), and conducted hardware generation on the searched network using the exhaustive-search tool in Section~\ref{sec:cascaded}.
It represents the typical separate design performed in practice.
Following \cite{proxylessnas}, the search was performed for 120 epochs with batch size of 256, while warmup (Section~\ref{sec:warmup}) was done for 40 epoches.
SGD optimizer with Nesterov momentum was used for the search using cosine scheduling with learning rate of 0.025, weight decay 0.00004 ($\lambda_1$), label smoothing 0.1 and momentum 0.9. 
After the search, the final network was trained from scratch for 300 epoches. 
The hyperparameters for training are the same, except that the learning rate is 0.008 and the weight decay factor is 0.001.
Also, we design EDD~\cite{edd} as the second baseline. Since EDD cannot be applied to hardware parameters for dataflow and register files, we conduct the co-exploration only based on the number of PEs and perform a post-search for the remaining parameters.
One problem we experienced with \cite{edd} under our setting is that it uses a loss function that multiplies classification loss with the latency loss as below:
\vspace{-1mm} 
\begin{equation}
    \mathcal{L}oss = \lambda_2 \cdot\mathcal{L}oss_{CE}\cdot \textstyle\sum{Latency}
    \vspace{-1mm}
\end{equation}
Note that $\lambda_2$ does not adjust the weight between the two terms. 
It renders a significant problem of the network shrinking too much to optimize latency quickly.
The result is that the solution results in a very low hardware cost, but shows an unacceptable accuracy as shown with EDD (Original) in Table~\ref{tbl:cifar}.
Thus, we also experiment with changing the loss function as Eq.~\ref{eq:overall} to alleviate the problem, which we denote as EDD + Proposed Loss func.

Using \nasname, we have performed co-exploration with the cost functions described in Section~\ref{sec:cost}.
For the $Cost_{HW\_linear}$, we set three different cost functions, which we named as latency-oriented, energy-oriented, and balanced.
All the remaining hyperparameters were the same as the baseline. 
Similar to after-search training, a one-time exact hardware generation was performed after the search to obtain the optimal hardware accelerator design.

Overall, \nasname is able to obtain network-accelerator designs superior to the baselines.
For comparison, we report two designs, one with high accuracy (-A) and the other towards efficient hardware design (-B).
For the high accuracy design (-A), \nasname achieves almost the same accuracy of the baseline (no penalty). 
For the efficient hardware design (-B), we chose the design with the best cost function within a 1\texttildelow2\% accuracy drop. 
It shows that \nasname achieves up to $10\times$ better EDAP, or $3\times$ better latency by performing an efficient co-exploration.
With the latency-oriented cost function, the latency becomes a much lower value than the others, while the energy-oriented cost function yields better energy consumption than the other two.
Overall, it shows that the user of \nasname can tune the cost hyperparameters to obtain a solution of interest.

\begin{figure}[t]
\scriptsize
\centering
    \includegraphics[width=0.7\columnwidth]{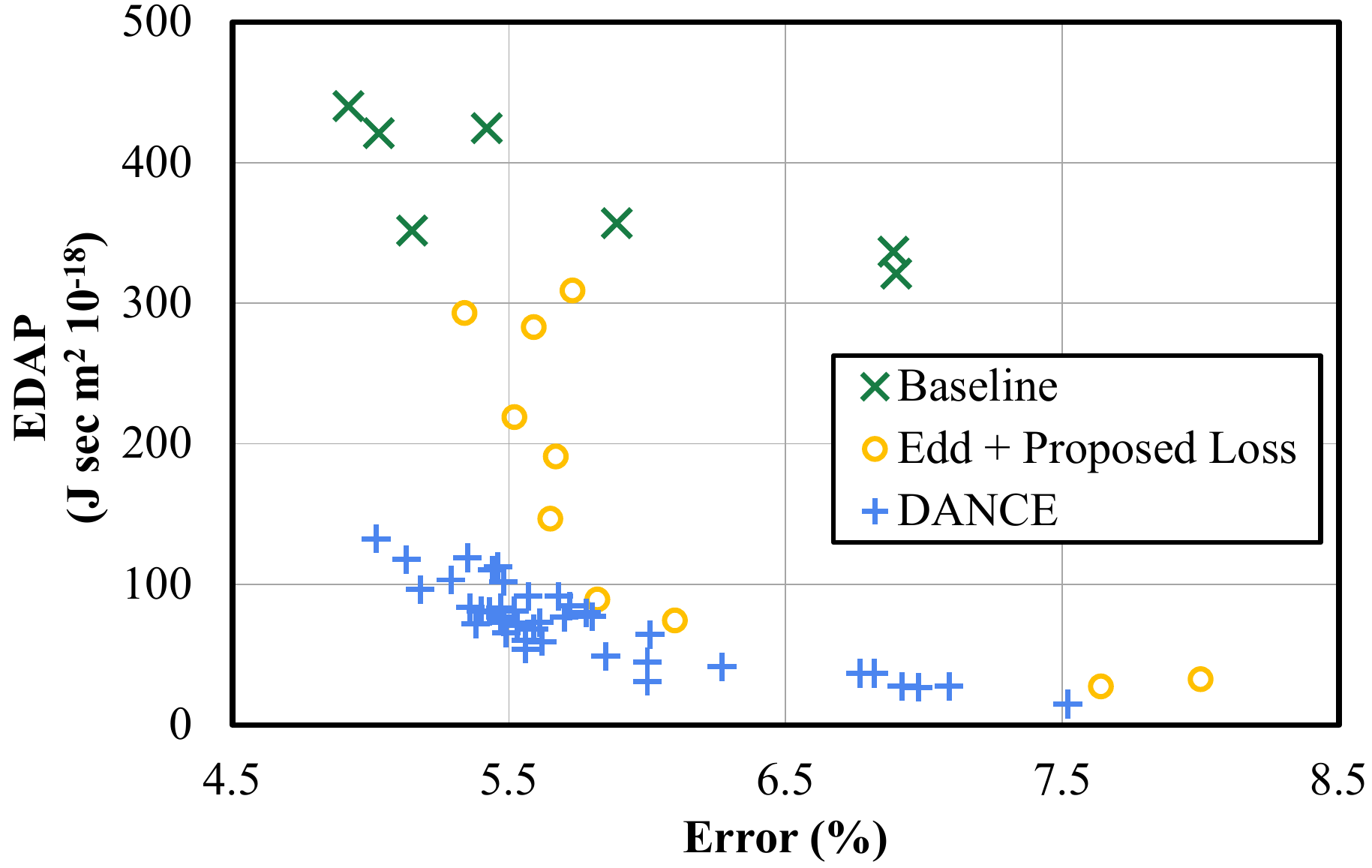}
\caption{Error-EDAP plot. Lower is better for both metrics.}\vspace{-3mm}
\label{fig:plot}
\end{figure}

\begin{figure*}[]

\centering
          
       \subcaptionbox{\scriptsize Searched network and accelerator design for a latency-oriented cost function.\label{fig:lat}}{
        \centering
        \includegraphics[width=0.48\textwidth]{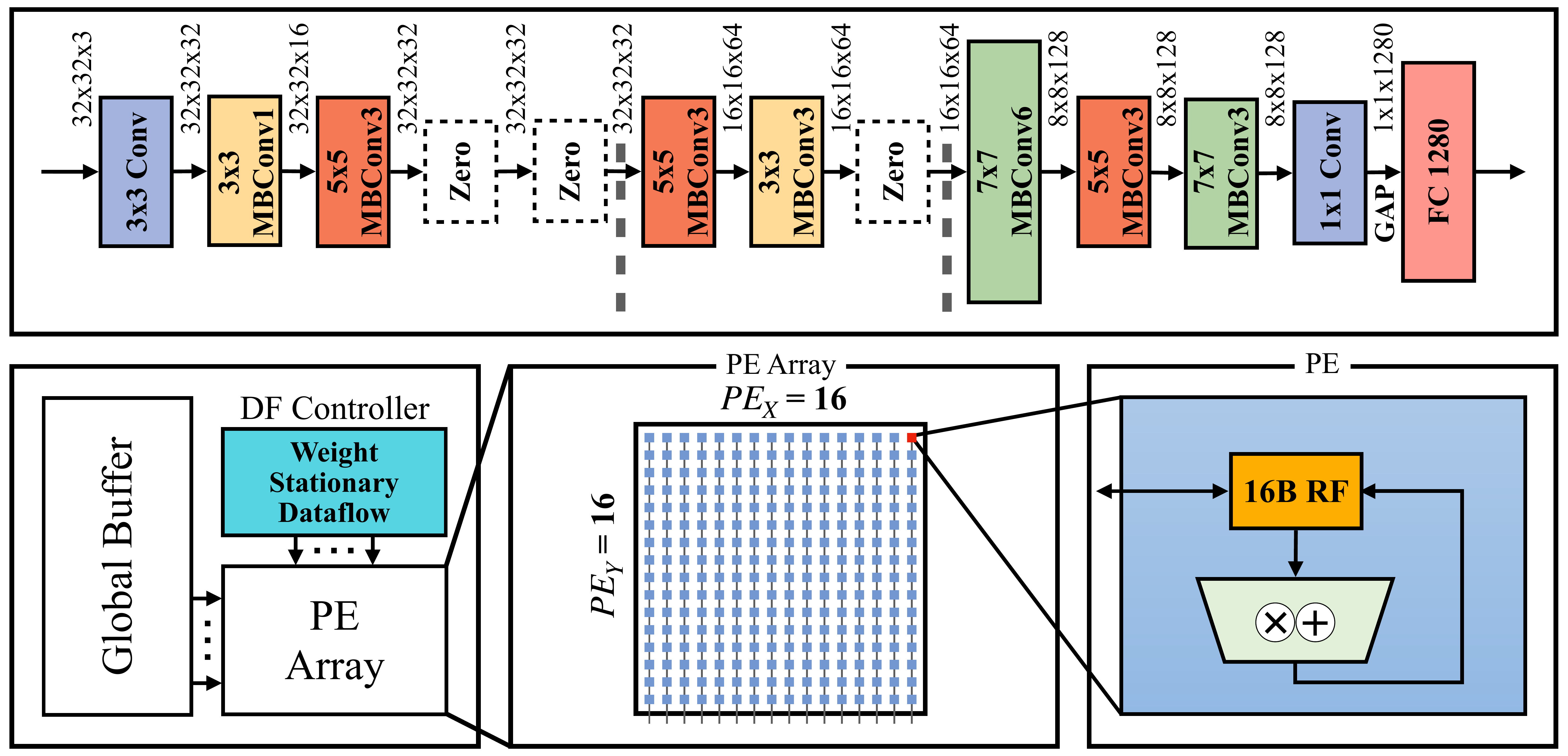}
    }
       \subcaptionbox{\scriptsize Searched network and accelerator design for an energy-oriented cost function.\label{fig:energy}}{
        \centering
        \includegraphics[width=0.48\textwidth]{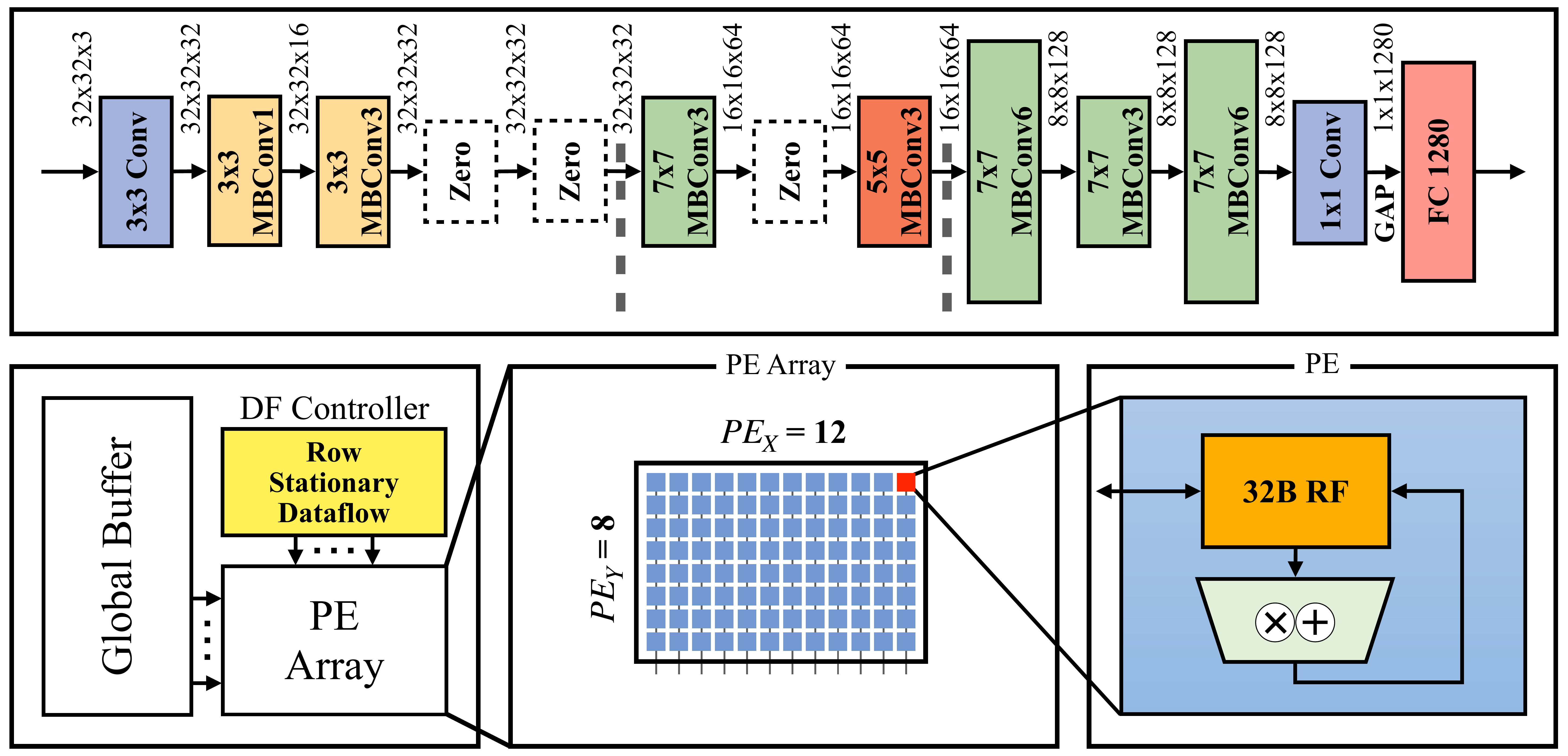}
    }
    
          \caption{Searched network and accelerator designs. Bold characters represent the design parameters.}\vspace{-4mm}
\label{fig:shape1}
\end{figure*}

\figurename~\ref{fig:plot} shows that \nasname searches dominating solutions compared to the baselines, not merely sacrificing accuracy with the hardware cost.
The figure plots the EDAP-error relations of the designs found from the baseline and \nasname.
In both of the axes, lower is better. 
We have searched with varying $\lambda_2$ from Eq.~\ref{eq:overall} to achieve different balance between accuracy and the $Cost_{HW}$.
The baselines and \nasname are both able to reach similar accuracies with accuracy-oriented hyperparameter settings, 
but \nasname shows a much better trade-off, and is able to gain superior cost metric than the baseline with Flops penalty.
Also, compared to EDD, we obtain superior performance of more than $2\times$ better EDAP under similar accuracies.
This is because EDD does not model the network-hardware relationship, and is unable to find efficient pair of design for the solution especially with high accuracies.
Note that EDD plotted in the figure uses the modified loss function with our proposal since the original EDD results in a too low accuracy.

\subsubsection{Experimental Results on ImageNet}
\tablename~\ref{tbl:imgnet} shows the performance of \nasname on ImageNet dataset.
The baseline with the separate hardware search yields 71.12\% accuracy but suffers from a large hardware cost. 
Applying flops penalty or EDD fails to find an efficient solution.
As in the case of experiments of \tablename~\ref{tbl:cifar}, \nasname discovers good tradeoff points, and results with significantly better cost metrics with only small accuracy drops, with up to 3$\times$ EDAP advantages. 

\begin{table}[b]
\vspace{-3mm}
    \centering
    \scriptsize
   \caption{Comparison of Existing Co-exploration Algorithms}
    \label{tbl:coex}
    {
\def\arraystretch{0.9}%
\setlength\tabcolsep{0.5pt}
    \begin{tabular}{p{9mm}ccccccc}
     \toprule
  Alg. &  Backbone &  Dataset &Acc.(\%) & GPU-hours & Candidates  &  Method & \makecell{Net-HW\\ Relation}\\        
    \midrule
        \cite{cosearch_fpga} & Custom & DAC-SDC & 68.6 & N/A & 68  & CD$^{*}$ & \cmark\\
        \cite{cosearch_fpga2} & Custom &  CIFAR-10 & 89.7 & N/A & N/A & RL & \cmark \\
        \cite{cosearch_multi} & ResNet-9 &  CIFAR-10 &  93.2 &  3.5h & \texttildelow160 & RL  & \cmark\\
        \cite{cosearch_bestofboth} & NASBench &  CIFAR-100 & 74.2 & 2300h & 2300 & RL  & \cmark  \\
         \cite{cosearch} & ProxylessNAS & CIFAR-10 & 85.2  & 103.9h & 308 & RL  & \cmark  \\
         \cite{edd}$^\dagger$ & ProxylessNAS & CIFAR-10 & 94.4  & 3h & 1 & gradient  & \xmark  \\
    \midrule
        \textbf{\nasname} & ProxylessNAS & CIFAR-10 & \textbf{ 95.0} & \textbf{3h} & \textbf 1 & \textbf{ gradient} & \cmark \\
        \bottomrule
        \multicolumn{8}{r}{\tiny $^\dagger$Reproduced and modified for the same setting \ \ \ *CD = Coordinate Descent} \\

    \end{tabular}
    }\\
   
\end{table}

\subsection{Network and accelerator design searched by \nasname}
\label{sec:shape}
\figurename~\ref{fig:shape1} shows the two sets of network architecture and the accelerator design that has been optimized along with the network. 
We chose two cost efficient designs (-B) generated with the latency-oriented cost function and energy-oriented cost function as they reveal useful insights on how the network architecture along with the accelerator design are found.
In the figure, the values written with bold characters represent the design parameters to be searched with \nasname.

The latency-oriented network (\figurename~\ref{fig:lat}) has relatively smaller kernel sizes (e.g., 3x3 MBConv instead of 7x7 MBConv) compared to the energy-oriented network.
On the other hand, it has more channels coming from larger expand ratios.
Regardless of the dataflow, accelerators are good at utilizing channel-level parallelism, so having more channels helps increase the number of active PEs at a time, resulting in a lower latency. 
To achieve low latency with such network, the searched accelerator has a relatively larger PE array to accelerate the speed. 
Finally, the chosen WS (Weight Stationary) dataflow is known to be generally good at achieving low latency. 

The energy-oriented network (\figurename~\ref{fig:energy}) has relatively larger kernel sizes (7x7 MBConv) with narrower channel width. 
Even though having larger kernel sizes often lowers PE utilization and increases the latency, having large number of unutilized PEs does not contribute much towards high energy. 
This is because dynamic energy consumption mainly depends on the number of MAC operations and data accesses. 
On the other hand, having narrower channel width often yields lower energy consumption, since it reduces the number of accesses for input/output activations. 
Comparing layers that have equal number of MAC operations with a small kernel/wider width and a layer with a large kernel/narrow width, the former has better latency due to the high PE utilization, and the latter has better energy consumption due to low data access.
The accelerator for the energy-oriented cost function has been searched to have RS dataflow, which is known to often exhibit good energy efficiency~\cite{eyeriss}. 
The PE array is small, to reduce the energy consumption. 
$PE_Y$ is especially small, since the depth-wise convolutions have only one output channel, and reducing $PE_Y$ for low energy is more advantageous than reducing $PE_X$.
Each PE has a larger RF compared to that of the latency-oriented design, because larger RF means less accesses to the GB (Global Buffer) and results in a low energy consumption.

\subsection{Comparison of \nasname with Existing Co-exploration Algorithms}
\tablename~\ref{tbl:coex} compares \nasname with other accelerator/network co-exploration algorithms. 
Since the environments are all different (ASIC vs FPGA, different technology node, different NAS backbone, etc), direct comparison between the reported values is not feasible.  
Even accuracy cannot be directly compared since it depends on the underlying NAS algorithm. 
However, it hints at the method's searching capability if the difference is huge, and thus we report the accuracy and the search cost from the published literature for a rough comparison purpose.

Most of the co-exploration algorithms utilize reinforcement learning, 
and they suffer from having to train many candidates during search.
As a result, many of them only output sub-optimal network architectures with inferior accuracy.

The search time also reveals the advantage of \nasname,
which is an order of magnitude faster compared to RL-based works.
The difference is small for \cite{cosearch_multi}, but this is due to the fact that its backbone architecture is based on a manually fine-tuned architecture~\cite{resnet9} where the model size is significantly small.
The `candidates' column is an attempt to provide a fair comparison of the search costs for considering such cases.
It reports the number of candidates each algorithm has to train during the search. 
RL-based co-exploration algorithms require around hundreds to thousands of candidates to be trained, while \nasname requires only one. 
\cite{edd} is differentiable, and our reproduction with the same NAS backbone gives a similar accuracy and search cost. 
However, as it is unable to reflect the network-hardware relation, its resulting co-exploration solution shows significantly less quality than \nasname as demonstrated in Section~\ref{sec:co-ex}.

\section{Conclusion}
In this paper, we propose a novel differentiable method of co-exploring hardware accelerator and network architecture together targeting both high accuracy and low cost metrics.
We model a neural net based hardware evaluator, to obtain efficient hardware design without compromising the accuracy at an extremely low search cost.
We believe this work would bring cost reductions to the co-exploration problem in many additional fields in the future, such as video or natural language processing.

\label{sec:conc}

\printbibliography

\end{document}